# An Automated System for Epilepsy Detection using EEG Brain Signals based on Deep Learning Approach

Ihsan Ullah[1], Muhammad Hussain[2,*], Emad-ul-Haq Qazi[2] and Hatim Aboalsamh[2]
[1]Insight Centre for Data Analytics, National University of Ireland, Galway, Ireland
[2]Visual Computing Lab, Department of Computer Science, College of Computer and Information Sciences, King Saud University, Riyadh, Saudi Arabia
{[1] Ihsan.ullah@insight-centre.org, [2,*]mhussain@ksu.edu.sa, [2]emadhaq@gmail.com, [2]hatm@ksu.edu.sa }

## Abstract:

Epilepsy is a neurological disorder and for its detection, encephalography (EEG) is a commonly used clinical approach. Manual inspection of EEG brain signals is a time-consuming and laborious process, which puts heavy burden on neurologists and affects their performance. Several automatic techniques have been proposed using traditional approaches to assist neurologists in detecting binary epilepsy scenarios e.g. seizure vs. non-seizure or normal vs. ictal. These methods do not perform well when classifying ternary case e.g. ictal vs. normal vs. inter-ictal; the maximum accuracy for this case by the state-of-the-art-methods is 97±1%. To overcome this problem, we propose a system based on deep learning, which is an ensemble of pyramidal one-dimensional convolutional neural network (P-1D-CNN) models. In a CNN model, the bottleneck is the large number of learnable parameters. P-1D-CNN works on the concept of refinement approach and it results in 60% fewer parameters compared to traditional CNN models. Further to overcome the limitations of small amount of data, we proposed augmentation schemes for learning P-1D-CNN model. In almost all the cases concerning epilepsy detection, the proposed system gives an accuracy of 99.1±0.9% on the University of Bonn dataset.

## 1. Introduction

Epilepsy is a neurological disorder affecting about fifty million people in the world (Megiddo et al., 2016). Electroencephalogram (EEG) is an effective and non-invasive technique commonly used for monitoring the brain activity and diagnosis of epilepsy. EEG readings are analyzed by neurologists to detect and categorize the patterns of the disease such as pre-ictal spikes and seizures. The visual examination is time-consuming and laborious; it takes many hours to examine one day data recording of a patient, and also it requires the services of an expert. As such, the analysis of the recordings of patients puts a heavy burden on neurologists and reduces their efficiency. These limitations have motivated efforts to design and develop automated systems to assist neurologists in classifying epileptic and non-epileptic EEG brain signals.

Recently, a lot of research work have been carried out to detect the epileptic and non-epileptic signals as a classification problem (Gardner et al., 2006; Meier et al., 2008; Mirowski et al., 2009; Sheb et al., 2010). From the machine learning (ML) point of view, recognition of epileptic and non-epileptic EEG signals is a challenging task. Usually, there is a small amount of epilepsy data available for training a classifier due to infrequently happening of seizures. Further, the presence of noise and artifacts in the data creates difficulty in learning the brain patterns associated with normal, ictal, and non-ictal cases. This difficulty increases further due to inconsistency in seizure morphology among patients (McShane, 2004). The existing automatic seizure detection techniques use traditional signal processing (SP) and ML techniques. Many of these techniques show good accuracy for one problem but fail in performing well for others e.g. they classify seizure vs. non-seizure case with good accuracy but show bad performance in case of normal vs. ictal vs. inter-ictal (T. Zhang et al., 2017). It is still a challenging problem due to three reasons, i) there does not exist a generalized model that can classify binary as well as ternary problem (i.e. normal vs. ictal vs. inter-ictal), ii) less available labeled data, and ii) low accuracy. To help and aid neurologists, we need a generalized automatic system that can show good performance even with fewer training samples (Andrzejak et al., 2001; Sharmila et al., 2016).

* Corresponding author

Researchers have proposed methods for the detection of seizures using features extracted from EEG signals by hand-engineered techniques. Some of the proposed methods use spectral (Tzallas et al., 2012) and temporal aspects of information from EEG signals (Shoeb, 2009). An EEG signal contains low-frequency features with long time-period and high-frequency features with short time period (Adeli et al., 2003) i.e. there is a kind of hierarchy among features. Deep learning (DL) is a state-of-the-art ML approach which automatically encodes hierarchy of features, which are not data dependent and are adapted to the data; it has shown promising results in my applications. Moreover, features extracted using the DL models have shown to be more discriminative and robust than hand-designed features (LeCun et al., 1995). In order to improve the accuracy in the classification of epileptic and non-epileptic EEG signals, we propose a method based on DL.

The recent emergence of DL techniques shows significant performance in several application areas. The variants of deep CNN i.e. 2D CNN such as AlexNet (Krizhevsky et al., 2012), VGG (Simonyan et al., 2014) etc. or 3D networks such as 3DCNN (Ji et al., 2013), C3D (Tran et al., 2015) etc. have shown good performance in many fields. Recently, 1D-CNN has been successfully used for text understanding, music generation, and other time series data (Cui et al., 2016; Ince et al., 2016; LeCun et al., 1998; X. Zhang et al., 2015). The end-to-end learning paradigm of DL approach avoids the selection of a proper combination of feature extractor and feature subset selector for extracting and selecting the most discriminative features that are to be classified by a suitable classifier (Andrzejak et al., 2001; Hussain et al., 2016; Sharmila et al., 2016; T. Zhang et al., 2017). Although the traditional approach is fast in training as compared to DL approach, it is far slower at test time and does not generalize well. Trained deep models can test a sample in a fraction of a second, and are suitable for real-time applications; the only bottleneck is the requirement of a large amount of data and its long training time. To overcome this problem, an augmentation scheme needs to be introduced that may help in using a small amount of available data in an optimal way for training a deep model.

As EEG is a 1D signal, as such we propose a pyramidal 1D-CNN (P-1D-CNN) model for detecting epilepsy, which involves far less number of learnable parameters. As the amount of available data is small, so for training P-1D-CNN, we propose two augmentation schemes. Using trained P-1D-CNN models as experts, we designed a system as an ensemble of P-1D-CNN models, which employs majority vote strategy to fuse the local decisions for detecting epilepsy. The proposed system takes an EEG signal, segment it with fixed-size sliding window, and pass each sub-signal to the corresponding P-1D-CNN model (Fig. 2) that process it and gives the local decision to the majority-voting module. In the end, the majority-voting module takes the final decision (Fig. 1). It outperforms the state-of-the-art techniques for different problems concerning epilepsy detection. The main contributions of this study are: 1) data augmentation schemes, 2) a system based on an ensemble of P-1D-CNN deep models for binary as well as ternary EEG signal classification, 3) a new approach for structuring deep 1D-CNN model and 4) thorough evaluation of the augmentation schemes and the deep models for detecting different epilepsy cases.

The rest of the paper is organized as follows: In Section 2, we present the literature review. Section 3 describes in detail the proposed system. Model selection, data augmentation schemes, and training of P-1D-CNN model are discussed in Section 4. Section 5 presents, discuss and compare the results we achieved. In the end, Section 6 concludes the paper.

## 2. Literature Review

The recognition of epileptic and non-epileptic EEG signals is a classification problem. It involves extraction of the discriminatory features from EEG signals and then performing classification. In the following paragraphs, we give an overview of the related state-of-the-art techniques, which use different feature extraction and classification methods for classification of epileptic and non-epileptic EEG signals.

Swami et al. (Swami et al., 2016) extracted hand-crafted features such as Shannon entropy, standard deviation, and energy. They employed the general regression neural network (GRNN) classifier to classify these features and achieved maximum accuracy, i.e., 100% and 99.18% for A-E (non-seizure vs. seizure) and AB-E (normal vs. seizure) cases, respectively on University of Bonn dataset. However, maximum accuracy for other cases like B-E, C-E, D-E, CD-E, and ABCD-E is 98.4 %. In another study, Guo et al. (Guo et al., 2010) achieved the accuracy of 97.77% for ABCD-E case on the same dataset. They used artificial neural network classifier (ANN) to classify the line length features that were extracted by using discrete wavelet transform (DWT). Nicolaou et al. (Nicolaou et al., 2012) extracted the permutation entropy feature from EEG signals. They employed support vector machine (SVM) as a classifier and achieved an accuracy of 93.55% for A-E case on the University of Bonn dataset. However, maximum accuracy for other cases such as B-E, C-E, D-E, and ABCD-E is 86.1 %. Gandhi et al. (Gandhi et al., 2011) extracted the entropy, standard deviation and energy features from EEG signals using DWT. They used SVM and probabilistic neural network (PNN) as a classifier and reported the maximum accuracy of 95.44% for ABCD-E case. Gotman et al. (J Gotman et al., 1979) used sharp wave and spike recognition technique. They further enhanced this technique in (J Gotman, 1982; Jean Gotman, 1999; Koffler et al., 1985; Qu et al., 1993). Shoeb et al. (Shoeb, 2009) used SVM classifier and adopted a patient-specific prediction methodology; the results indicate that a 96% accuracy was achieved. In most of the works, common classifier used to distinguish between seizure and non-seizure events is support vector machine (SVM). However, in (Khan et al., 2012) linear discriminant analysis (LDA) classifier was used for classification of five subjects consisting of sixty-five seizures. It achieved 91.8%, 83.6% and 100 % accuracy, sensitivity, and specificity, respectively. Acharya et al. (Acharya et al., 2012) focused on using entropies for EEG seizure detection and seven different classifiers. The best-performing classifier was the Fuzzy Sugeno classifier, which achieved 99.4% sensitivity, 100% specificity, and 98.1% overall accuracy. The worst performing classifier was the Naive Bayes Classifier, which achieved 94.4% sensitivity, 97.8% specificity, and 88.1% accuracy. Nasehi and Pourghassem (Nasehi et al., 2013) used Particle Swarm Optimization Neural Network (PSONN), which gave 98% sensitivity. Yuan et al. (Yuan et al., 2012) used extreme learning machine (ELM) algorithm for classification. Twenty-one (21) seizure records were used to train the classifier and sixty-five (65) for testing. The results showed that the system achieved on average 91.92% sensitivity, 94.89% specificity and 94.9% overall accuracy. Patel et al. (Patel et al., 2009) proposed a low-power, real-time classification algorithm, for detecting seizures in ambulatory EEG. They compared Mahalanobis discriminant analysis (MDA), quadratic discriminant analysis (QDA), linear discriminant analysis (LDA) and SVM classifiers on thirteen (13) subjects. The results indicate that the LDA show the best results when it is trained and tested on a single patient. It gave 94.2% sensitivity, 77.9% specificity, and 87.7% overall accuracy. When generalized across all subjects, it gave 90.9% sensitivity, 59.5% specificity, and 76.5% overall accuracy. Further, a detailed list of feature extractors and classifiers used for binary (e.g. epileptic vs. non-epileptic) and ternary (ictal vs. normal vs. interictal) scenarios is given in (Sharmila et al., 2016; T. Zhang et al., 2017).

The overview of the state-of-the-art given above indicates that most of the feature extraction techniques are hand-crafted, which are not adapted to the data. In order to improve the accuracy and generalization of an epilepsy detection system, DL approach can be used to avoid the need for hand-crafted feature extractors and classifiers. To the best of our knowledge, so far no one used DL approach for epilepsy detection, perhaps the reason is the small amount of available data, which is not enough to train a deep model. As such, we felt motivated to employ DL technique for proposing a deep model that involves a small number of learnable parameters and classifies efficiently EEG brain signals as epileptic or non-epileptic.

## 3. The Proposed System

The proposed system for automatic epilepsy detection using EEG brain signals based on deep learning is shown in Fig. 1. It consists of three main modules: (i) splitting the input signal into sub-signals using a fixed-size overlapping windows, (ii) an ensemble of P-1D-CNN models, where each sub-signal is classified by the corresponding P-1D-CNN model, and (iii) fusion and decision, the local decisions are fused using majority vote to take the final decision.

A general deep model needs a huge amount of data for training, but for epilepsy detection problem the amount of data is limited. To tackle this issue, we introduce data augmentation schemes in Section 4, where each EEG signal corresponding to epilepsy or normal case is divided into overlapping windows (sub-signals) and each window is treated as an independent instance to train P-1D-CNN model. Using copies of the trained P-1D-CNN model, we build ensemble classifier, where each model plays the role of an expert examining a certain part of the signal. For classification, keeping in view the augmentation approach, an input EEG signal is split into overlapping windows, which are passed to different P-1D-CNN models in the ensemble, as shown in Fig. 1, i.e. different parts of the signal are assigned to different experts (models) for its local analysis. After local analysis, each model provides a local decision; lastly, these decisions are fused using majority vote for final decision. The number of P-1D-CNN models (experts) in the ensemble depends on the number of windows. For example, in case an input EEG signal is divided into $n$ windows (sub-signals), the ensemble will consist of $n$ P-1D-CNN models.

The core component of the system is a P-1D-CNN model. It is a deep model, which consists of convolutional, batch normalization, ReLU, fully connected and dropout layers. In the following section, we present the detail of this deep model. For compactly describing the ideas, key terms and their acronyms are given in Table 1.

**Table 1.** Key terms and their acronyms that will be used throughout the paper

| Name | Abbreviation | Name | Abbreviation |
|---|---|---|---|
| Accuracy | $Acc$ | No of Kernels | $K$ |
| Accuracy with Voting | $Acc_V$ | Size of Receptive field | $Rf$ |
| Fully connected | $FC$ | Batch Normalization | $BN$ |
| Rectifier Linear activation Unit | $ReLU$ | Drop out | $DO$ |
| Specificity | $Spe$ | Sensitivity | $Sen$ |
| Geometric Mean | $G_M$ | F-Measure | $F_M$ |
| 10-fold Validation | K-number | Standard Deviation | $std$ |

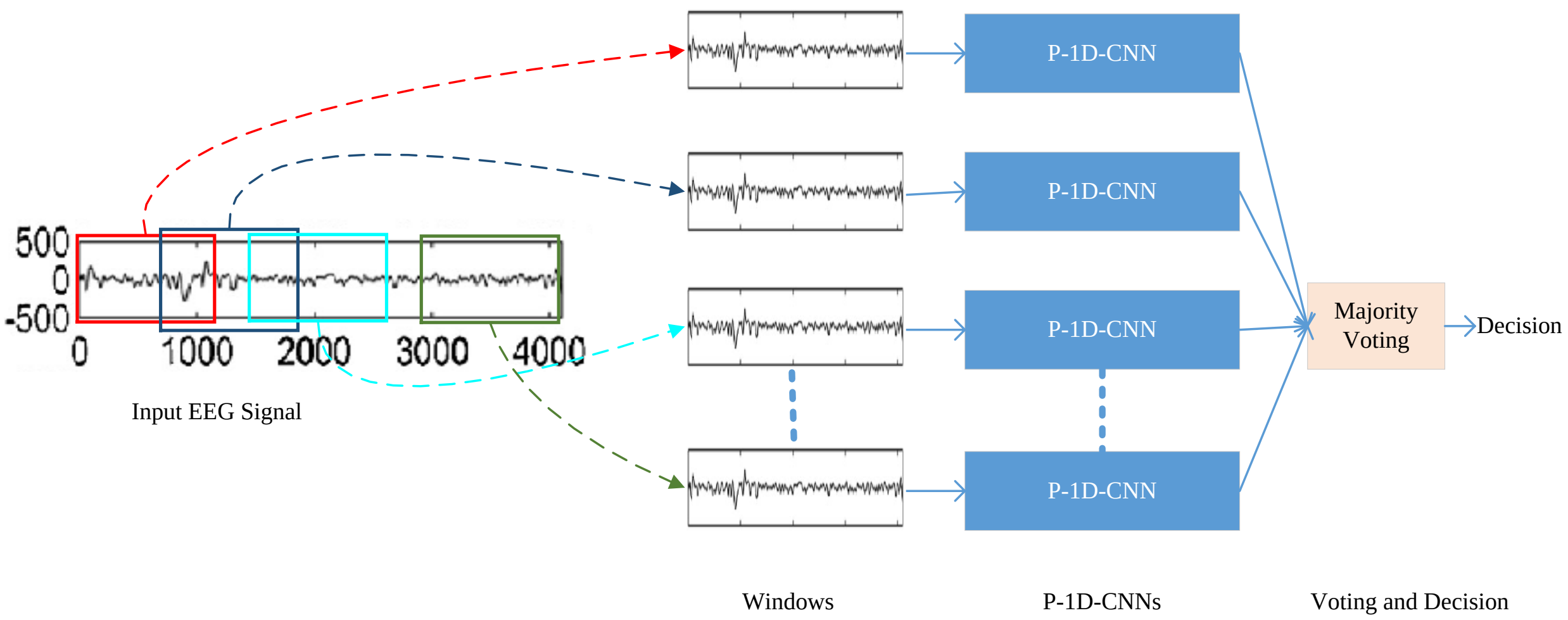

**Fig. 1.** Overall structure of our automatic ensemble deep EEG classification

### 3.1 P-1D-CNN Architecture

A deep CNN model (LeCun et al., 1998; Simonyan et al., 2014) learns structures of EEG signals from data automatically and performs classification in an end-to-end manner, which is opposite to the traditional hand-engineered approach, where first features are extracted, a subset of extracted features are selected and finally passed to a classifier for classification. The main component of a CNN model is a convolutional layer consisting of many channels (feature maps). The output of each neuron in a channel is the outcome of a convolution operation with a kernel (which is shared by all neurons in the same channel) of fixed receptive field on input signal or feature maps (1D signals) of the previous convolutional layer. In this way, CNN analyses a signal to learn a hierarchy of discriminative information. In CNN, the kernels are learned from data unlike hand-engineered approach, where kernels are predefined e.g. wavelet transform. Although CNN with its novel idea of shared kernels has the advantage of a significant reduction in the number parameters over fully connected models, the recent emergence of making CNN deeper has given rise to a very large number of parameters adding to its complexity resulting in overfitting over a small dataset. As available EEG data for epilepsy detection is small in size, we handled this problem using two different strategies i.e. novel data augmentation schemes and a memory efficient deep CNN model with a small number of parameters.

EEG signal is a 1D time series; as such for its analysis, we propose pyramidal 1D-CNN model, which we call P-1D-CNN and its generic architecture is shown in Fig. 2, it is an end-to-end model. Unlike traditional CNN models, it does not include any pooling layer; the redundant or unnecessary features are reduced with the help of bigger strides in convolution layers. Convolutional and fully connected layers learn a hierarchy of low to high-level features from the given input signal. The high-level features with semantic representation are passed as input to the softmax classifier in the last layer to predict the respective class of the input EEG signal.

A CNN model is commonly structured by adopting course to fine approach, where low-level layers have a small number of kernels, and high-level layers contain a large number of kernels. But this structure involves a huge number of learnable parameters i.e. its complexity is high. In stead, we adopted a pyramid architecture similar to the one proposed by Ullah and Petrosino (Ullah et al., 2016) for deep 2D CNN, where low-level layers have a large number of kernels and higher level layers contain small number of kernels.

This structure significantly reduces the number of learnable parameters, avoiding the risk of overfitting. A large number of kernels are taken in a Conv1 layer, which are reduced by a constant number in Conv2 and Conv3 layers e.g. Models E and H, specified in Table 3, contain Conv1, Conv2 and Conv3 layers with 24, 16, and 8 kernels, respectively. The idea is that low-level layers extract a large number of microstructures, which are composed of higher level layers into higher level features, which are small in number but discriminative, as the network gets deeper.

To show the effectiveness of the pyramid CNN model, we considered eight models, four of which have pyramid architecture. Table 3 shows detailed specifications of these models and also gives the number of parameters to be trained in each model. The last fully connected layer has two or three nodes depending on whether the EEG brain signal classification problem is two class (e.g. epileptic and non-epileptic ) or three class (normal vs. ictal vs. interictal). With the help of these models, we show how a properly designed model can result in equal or better performance despite fewer parameters, which has less risk of overfitting. The models having pyramid architecture involve significantly less number of learnable parameters, see Table 3; Model H, which has pyramid architecture, has 63.64% fewer parameters than D, a similar 1D-CNN model.

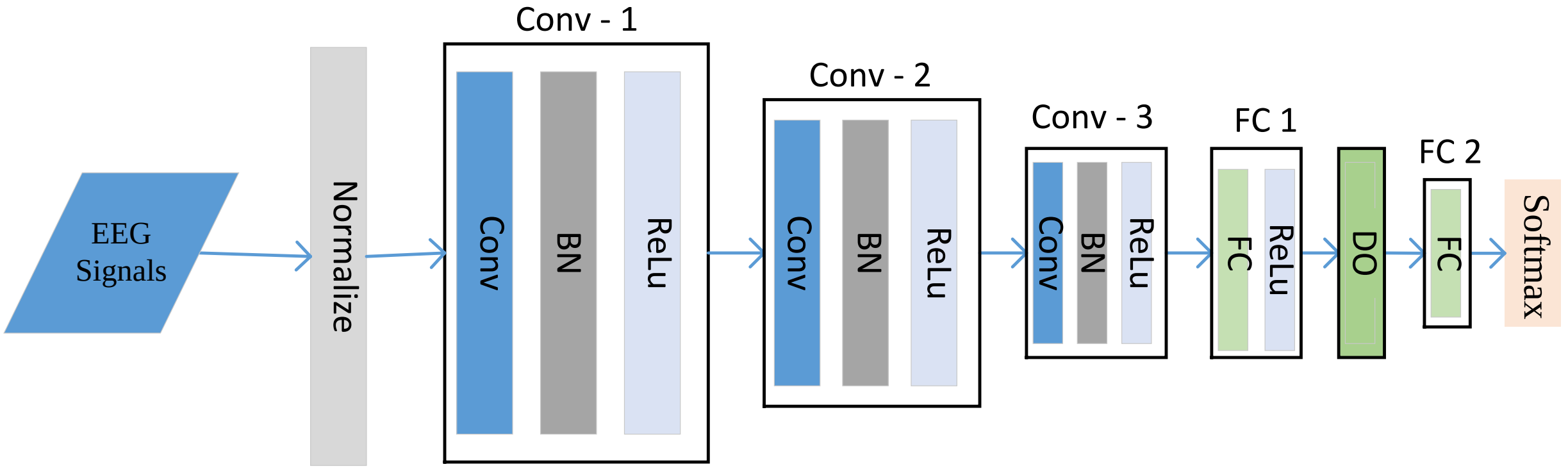

**Fig. 2.** The proposed Deep Pyramidal 1D-CNN Architecture (P-1D-CNN).

The detail of deep P-1D-CNN model is shown in Fig. 2. The input signals are normalized with zero mean and unit variance. This normalization helps in faster convergence and avoiding local minima. The normalized input is processed by three convolutional blocks, where each block consists of three layers: Convolutional layer ($Conv$), Batch normalization layer ($BN$) and non-linear activation layer ($ReLU$). The output of $ReLU$ layer in the third block is passed to a fully connected layer ($FC1$) that is followed by a $ReLU$ layer and another fully connected layer ($FC2$). In order to avoid overfitting, we use dropout before $FC2$. The output of $FC2$ is given to a softmax layer, which serves as a classifier and predicts the class of the input signal. The number of neurons in the final layer will change according to the number of classes to classify e.g. normal vs ictal vs inter-ictal (three classes) or Non-Seizure vs Seizure (Binary Class) shown by 2/3 in Table. 3. In the following subsections, we will briefly explain two of the main layers i.e. 1D-Convolutional and BN layers.

### a) Convolution Layers

The 1D-convolution operation is commonly used to filter 1D signals (e.g. time series) for extracting discriminative features. A convolutional layer is generated by convolving the previous layer with $K$ kernels of receptive field $Rf$ and depth, which is equal to the number of channels or feature maps in the previous layer. Formally, convolving the layer X = $\{x_{ij}: 1 \leq i \leq c, 1 \leq j \leq z\}$, where c is the number of channels in the layer and $z$ is the number of units in each channel, with $K$ kernels $k^i, l = 1, 2, \cdots, K$ each of receptive field $Rf$ and depth c yield the convolutional layer Y= $\{y_{ij}: 1 \leq i \leq m, 1 \leq j \leq K\}$, where

$$y_{ij} = \sum_{d=1}^{c} \sum_{e=1}^{k} w_{d,e}\, x_{i+d,j+e}, \qquad (1)$$

and $m$ is the number of units in each channel of the layer. Note that the number of channels in the generated convolutional layer is equal to the number of kernels. Different kernels extract different types of discriminative features from the input signal. The number of kernels varies as the network goes deeper and deeper. The low-level layer kernels learn micro-structures whereas the higher level layer kernels learn high-level features. In the proposed model, maximum number of kernels are selected in first convolution layer that is reduced by 33% in subsequent layers to maintain a pyramid structure. The activations (channels) of three convolutional layers are shown in Fig. 3.

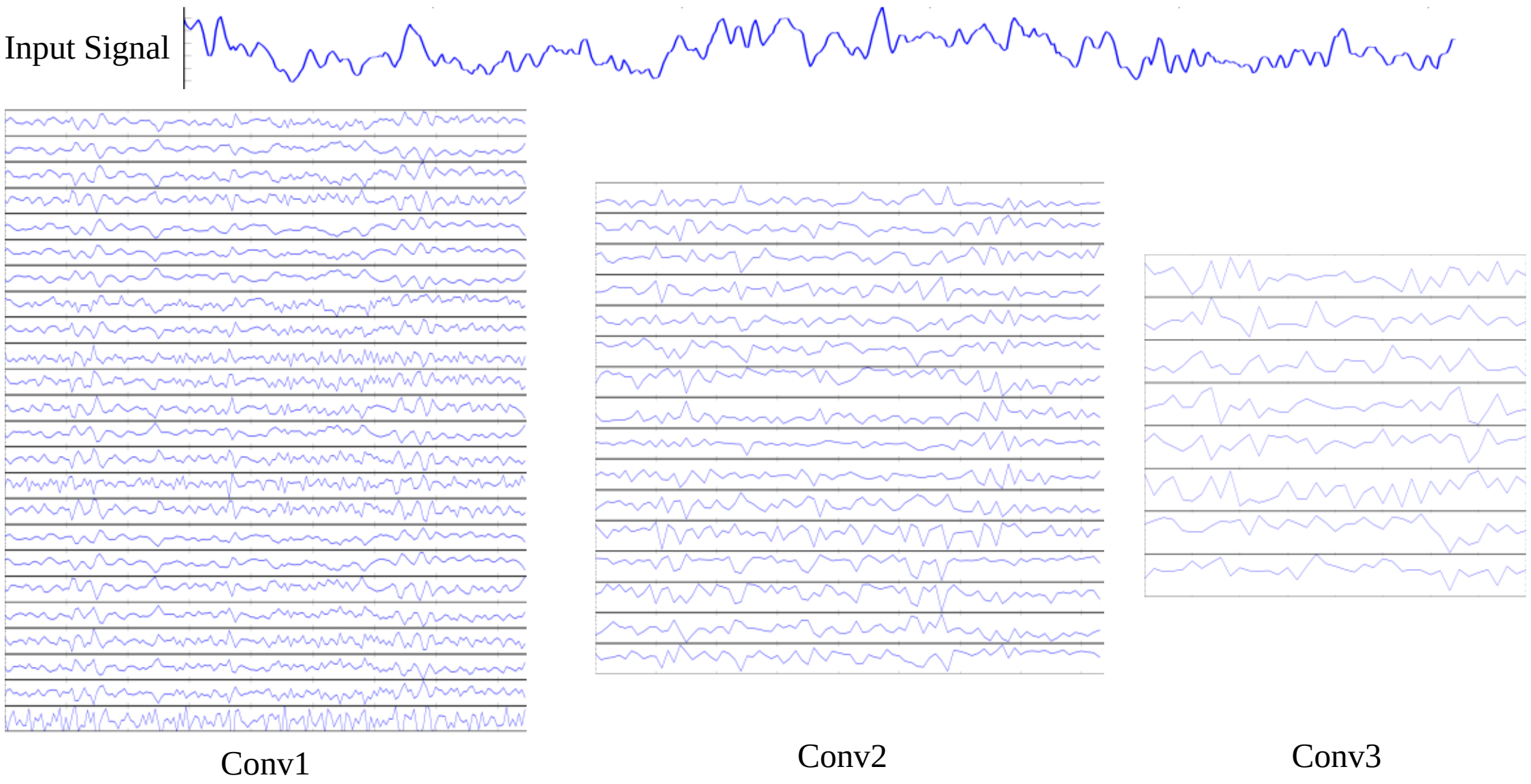

**Fig. 3.** Input Signal (first row), and activations of $Conv1$ (24 channels), $Conv2$ (16 channels) and $Conv3$ (08 channels) of P-1D-CNN model.

### b) Batch Normalization

During training, the distribution of feature maps changes due to the update of parameters, which forces to choose small learning rate and careful parameter initialization. It slows down the learning and makes the learning harder with saturating nonlinearities. Ioffe and Szegedy (Ioffe et al., 2015) called this phenomenon *as* internal covariate shift and proposed batch normalization (BN) as a solution to this problem. In BN, the activations of each mini batch at each layer are normalized, the detail can be found in (Ioffe et al., 2015). It is now very common to use BN in neural networks. It helps in avoiding special initialization of parameters, yet provides faster convergence. In the proposed model, we use $BN$ after every convolutional layer.

## 4. Model Selection and Parameter Tuning

First, we present the detail of data, and the proposed data augmentation schemes. Then, we give evaluation measures, which have been used to validate the performance of the proposed system. After this, the training procedure has been elaborated. Finally, the best data augmentation scheme and P-1D-CNN model have been suggested by analyzing the results with different ways of data augmentation, and different 1D-CNN models.

### 4.1. Dataset and Data Augmentation Schemes

The data set used in this work was acquired by a research team at University of Bonn (Andrzejak et al., 2001) and have been extensively used for research on epilepsy detection. The EEG signals were recorded using standard 10-20 electrode placement system. The complete data consists of five sets (A to E), each containing 100 one-channel instances. Sets A and B consists of EEG signals recorded from five healthy volunteers while they were in a relaxed and awake state with eyes opened (A) and eyes closed (B), respectively. Sets C, D, and E were recorded from five patients. EEG signals in set D were taken from the epileptogenic zone. Set C was recorded from the hippocampal formation of opposite hemisphere of the brain. Sets C and D consist of EEG signals measured during seizure-free intervals (interictal), whereas, the EEG signals in Set E were recorded only during seizure activity (ictal) (Andrzejak et al., 2001). The detail is given in Table 2.

**Table 2.** University of Bonn epilepsy dataset details

| A | B | C | D | E |
|---|---|---|---|---|
| Non-Epileptic | Non-Epileptic | Epileptic | Epileptic | Epileptic |
| Eyes opened | Eyes closed | Interictal | Interictal | Ictal |

The number of instances collected in this dataset are not enough to train a deep model. Acquiring a large number of EEG signals for this problem is not practical and their labeling by expert neurologists is not an easy task. We need an augmentation scheme that can help us in increasing the amount of the data that is enough for training deep CNN model, which requires large training data for better generalization. The available EEG data is small that can learn the model but overfitting is evident. To overcome this problem, we propose two data augmentation schemes for training our model.

Each record in the dataset consists of 4097 samples. For generating many instances from one record, we adopted the sliding window approach similar to the one presented in refs. (Sharmila et al., 2016; T. Zhang et al., 2017). In (T. Zhang et al., 2017), Zhang et al. adopted a window size of 512 with a stride of 480 (93.75% of 512); each record is segmented into 8 equal EEG sub-signals, discarding the last samples. In this way, a total of 800 data instances are obtained for each dataset from 100 single-channel records, but this amount is not enough for learning the deep model. However, this approach indicates that the large stride is not helpful and smaller strides can be used for creating enough data. Based on the window size and stride, we propose two data augmentation schemes.

#### *Scheme-1*

The available signals are divided into disjoint training and testing sets, which consist of 90% and 10% of total signals, respectively. Data is augmented using training set. Choosing a window size of 512 and a stride of 64 (12.5% of 512 with an overlap of 87.5%), each signal of length 4097 in the training set is divided into 57 sub-signals, each of which is treated as an independent signal instance $S^{tr}$. In this way, a total of 5130 instances are created for each category (class), which are used to train the P-ID-CNN model. The *n* instances of the trained P-ID-CNN model are used to from an ensemble.

For testing, each signal of length 4097 in the testing set is divided into 4 sub-signals $S^{ts}$, each of length 1024; these sub-signals are treated as independent signal instances for testing. Each signal instance $S^{ts}$of length 1024 is divided further into three sub-signals with a window of size 512 and 50% overlap. This gives rise to 3 independent signal instances $S_i^{ts}, i = 1, 2, 3$, each of size 512, which are passed to three trained P-ID-CNN models in the ensemble and majority vote is used as a fusion strategy to take the decision about the signal instance $S^{ts}$. Each model in the ensemble plays the role of an expert, which analysis a local part of the signal instance $S^{ts}$ independently and the global decision is given by ensemble by fusing the local decisions.

### *Scheme-2*

This method is similar to scheme-1. In this case, the window size is 512 with an overlap of 25% (i.e. stride of 128) for creating training instances $S^{tr}$. For testing, each testing signal instance $S^{ts}$ of length 1024 is divided into three sub-signals with a window of size 512 and 75% overlap. This gives rise to 5 independent signal instances $S_i^{ts}, i = 1, 2, 3, 4, 5$, each of size 512, which are passed to five trained P-ID-CNN models in the ensemble and majority vote is used as a fusion strategy to take the decision about the signal instance $S^{ts}$.

### 4.2. Performance Measures (Evaluation procedure)

For evaluation, we adopted 10-fold cross validation for ensuring that the system is tested over different variations of data. The 100 signals for each class divided into 10 folds, each fold (10%), in turn, is kept for testing while the remaining 9 folds (90% signals) are used for learning the model. The average performance is calculated for 10 folds. The performance was evaluated using well-known performance metrics such as accuracy, specificity, sensitivity, precision, f-measure, and g-mean. Most of the state-of-the-art systems for epilepsy also employ these metrics, the adaptation of these metrics for evaluating our system helps in fair comparison with state-of-the-art systems. The definitions of these metrics are given below

$$Accuracy\ (Acc) = \frac{TP + TN}{Total\ Samples} \tag{2}$$

$$Specificity\ (Spe) = \frac{TN}{TN + FP} \tag{3}$$

$$Sensitivity\ (Sen) = \frac{TP}{FN + TP} \tag{4}$$

$$F - Measure\ (F_M)) = \frac{2 * Precision * Sensitivity}{Precision + Sensitivity} \tag{5}$$

$$G - Mean\ (G_M)\ = \sqrt{Specificity * Sensitivity} \tag{6}$$

where *TP* (true positives) is the number of abnormal cases (e.g. epileptic), which are predicted as abnormal, *FN* (false negatives) is the number of abnormal cases, which are predicted as normal, *TN* (true negatives) is the number of normal case that is predicted as normal and *FP* (false positives) is the number of normal cases that are identified as abnormal by the system.

#### 4.2.1 Training P-1D-CNN

Training of P-1D-CNN needs the weight parameters (kernels) to be learned from the data. For learning these parameters, we used the traditional back-propagation technique with cross entropy loss function and stochastic gradient descent approach with Adam optimizer (Kingma et al., 2014). Adam algorithm has six hyper-parameters: learning rate (0.001), beta1 (0.9), beta2 (0.999), epsilon (0.00000001), use locking (false) and name (Adam); we used default values of all these parameters (given in parentheses) except learning rate, which we set to a very small number of 0.00002. Although BN normally allows higher learning rate, a small learning rate is needed to control the oscillation of the network and to avoid any local minima problem when using Adam optimizer. The model is trained with a different number of iterations depending on the size of the dataset. In dropout, a probability value of 0.5 is used in all the experiments. The model was implemented in TensorFlow ("TensorFlow, 2017,"), a freely available DL library from Google. The number of iterations varies for each experiment – depending on the number of datasets we are using at one time in that experiment. For example, if we are using two datasets i.e. A vs E or D vs E, we trained the model with 50k iterations; if we are using three sets among five (i.e. A, B, C, D, or E) in an experiment e.g. AB vs C, we set maximum iterations to 150k. Whereas, if we are using four or all of the five available signal sets, we train our model with 300k iterations. Although the model trains much faster, still we train it to a maximum number of assigned iterations for better generalization of the model.

#### 4.2.2 Selection of Best Model and Data Augmentation Scheme

For selecting the best model, we considered eight CNN models in our initial experiments, as is shown in Table 3. For best model selection, we need to address two questions: a) which data augmentation scheme is the most suitable one? b) does pyramid-like structure have better generalization than the traditional model, where the number of kernels increases as the network goes deeper and deeper? To answer these questions, we performed in-depth experiments using 10-fold cross validation with all the eight models only on three class problem: non-epileptic (AB) vs epileptic inter-ictal (CD) vs epileptic ictal (E), which is the most challenging problem. These experiments led us to select the best model and the data augmentation scheme, which we used for other classification problems. It should be noted that all the 10-fold cross-validation sets are created randomly forcing to include all samples in training (90%) and testing (10%).

**Table 3.** The specifications of 8 1D-CNN models and their mean performance using 10-fold cross-validation for the AB vs. CD vs. E case.

| Model | | $M1$ | $M2$ | $M3$ | $M4$ | $M5$ | $M6$ | $M7$ | $M8$ | |
|---|---|---|---|---|---|---|---|---|---|---|
| $Conv1$ | $K$ | 8 | 8 | 8 | 8 | 24 | 24 | 24 | 24 | |
| | $Rf$ | 5 | 5 | 5 | 5 | 5 | 5 | 5 | 5 | |
| | $St$ | 3 | 3 | 3 | 3 | 3 | 3 | 3 | 3 | |
| $BN$ | - | - | - | - | - | - | - | - | - | |
| $ReLU$ | - | - | - | - | - | - | - | - | - | |
| $Conv2$ | $K$ | 16 | 16 | 16 | 16 | 16 | 16 | 16 | 16 | |
| | $Rf$ | 3 | 3 | 3 | 3 | 3 | 3 | 3 | 3 | |
| | $St$ | 2 | 2 | 2 | 2 | 2 | 2 | 2 | 2 | |
| $BN$ | - | - | - | - | - | - | - | - | - | |
| $ReLU$ | - | - | - | - | - | - | - | - | - | |
| $Conv3$ | $K$ | 24 | 24 | 24 | 24 | 8 | 8 | 8 | 8 | |
| | $Rf$ | 3 | 3 | 3 | 3 | 3 | 3 | 3 | 3 | |
| | $St$ | 2 | 2 | 2 | 2 | 2 | 2 | 2 | 2 | |
| $BN$ | - | - | - | - | - | - | - | - | - | |
| $ReLU$ | - | - | - | - | - | - | - | - | - | |
| $FC1$ | - | 20 | 20 | 40 | 40 | 20 | 20 | 40 | 40 | |
| $DO$ | - | 0 | 0.5 | 0 | 0.5 | 0 | 0.5 | 0 | 0.5 | |
| $FC2$ (Out) | - | 2/3 | 2/3 | 2/3 | 2/3 | 2/3 | 2/3 | 2/3 | 2/3 | |
| Parameters | | 21366/21387 | | 41106/41147 | | **8326/8347** | | **14946/14987** | | |
| | | | | | | | | | | Avg±std |
| AB vs CD vs E Aug. Scheme-1 | $Acc$ | 96.23 | 96.00 | 96.03 | 95.92 | 96.27 | 96.18 | 96.12 | 96.45 | 96.45±0.13 |
| | $Acc_V$ | 99.10 | 98.95 | 98.95 | 99.15 | 99.10 | 98.95 | 99.05 | 98.75 | 99.00±0.08 |
| | $std$ | 0.02 | 0.01 | 0.01 | 0.01 | 0.02 | 0.01 | 0.02 | 0.02 | |
| | $Sen$ | 0.96 | 0.96 | 0.96 | 0.96 | 0.95 | 0.96 | 0.96 | 0.97 | 0.960±0.003 |
| | $Spe$ | 0.98 | 0.98 | 0.97 | 0.98 | 0.98 | 0.97 | 0.98 | 0.96 | 0.975±0.004 |
| | $G-M$ | 0.97 | 0.97 | 0.97 | 0.97 | 0.97 | 0.97 | 0.97 | 0.96 | 0.968±0.000 |
| | $F-M$ | 0.95 | 0.96 | 0.95 | 0.96 | 0.96 | 0.96 | 0.96 | 0.95 | 0.956±0.004 |
| | | | | | | | | | | |
| AB vs CD vs E Aug. Scheme-2 | $Acc$ | 94.88 | 95.78 | 95.10 | 95.55 | 94.95 | 95.67 | 95.28 | 96.00 | 95.40±0.35 |
| | $Acc_V$ | 98.85 | 98.90 | 99.00 | 98.85 | 99.00 | 99.05 | 98.85 | 98.95 | 98.93±0.08 |
| | $std$ | 0.02 | 0.01 | 0.02 | 0.02 | 0.01 | 0.02 | 0.02 | 0.02 | |
| | $Sen$ | 0.95 | 0.97 | 0.95 | 0.96 | 0.96 | 0.96 | 0.95 | 0.97 | 0.958±0.007 |
| | $Spe$ | 0.97 | 0.97 | 0.98 | 0.98 | 0.97 | 0.97 | 0.98 | 0.97 | 0.973±0.005 |
| | $G-M$ | 0.96 | 0.97 | 0.96 | 0.97 | 0.96 | 0.97 | 0.96 | 0.97 | 0.965±0.005 |
| | $F-M$ | 0.96 | 0.97 | 0.96 | 0.97 | 0.96 | 0.97 | 0.96 | 0.97 | 0.965±0.005 |

The models were trained and tested using data augmentation schemes 1 and 2. Models $M1$ to $M4$ are designed using the traditional concept of increasing $K$ (the number of filters or kernels) in each higher layer as the network goes deeper, whereas models $M5$ to $M8$ (pyramid models) are designed using the concept of course to fine refinement approach i.e. reduce $K$ (the number of filters or kernels) by ratio of 33% in this case as the network goes deeper. The pyramid models involve a fewer number of parameters than traditional models, and as such are less prone to overfitting and generalize well.

The average performance results obtained using 10-fold cross-validation of different models and the data augmentation schemes are given in Table 3. First, the average accuracies (over all models) along with their standard deviations are 96.45±0.13 and 95.40±0.35 using data augmentation schemes 1 and 2, respectively; almost similar results can be observed in terms of other performance measures. It indicates that augmentation scheme 1 results in better performance than scheme 2. Based on this observation, scheme 1 is adopted for all other experiments in the paper.

Secondly, based on overall results it can be observed that pyramid model ($M5$ to $M8$) show results, which are better than or equal to those by traditional models with both augmentation schemes. Further, in most of the cases, the best result is given by pyramid model $M5$ with dropout 0.5 and 20 neurons in the fully connected layer; it works better with 20 neurons rather than 40 in the fully connected layer. It is obvious that *M5* is the optimal model, its gives slightly higher or similar performance but involves the minimum number parameters among all; such a model is easy to deploy on low-cost chips with limited memory as compared to the models with more parameters ($M1 - M4$). In all onward experiments, we will use model $M5$ with augmentation scheme 1.

## 5. Results and Discussion

After model selection, i.e. *M5* with augmentation scheme 1, we present and discuss the results for different experiment cases related to epilepsy detection. We considered three experiment cases: (i) normal vs inter-ictal vs ictal (AB vs CD vs E), (ii) normal vs epileptic (AB vs CDE and AB vs CD), (iii) seizure vs non-seizure (A vs E, B vs E, A+B vs E, C vs E, D vs E, C+D vs E). The comparison is done with state-of-the-art on 16 experiments: AB vs. CD vs. E, AB vs. CD, AB vs. E, A vs. E, B vs. E, CD vs. E, C vs. E, D vs. E, BCD vs. E, BC vs. E, BD vs. E, AC vs. E, ABCD vs. E, AB vs. CDE, ABC vs. E and ACD vs. E. Among 16 experiments, 14 have been frequently considered in most of the studies e.g. (Sharmila et al., 2016). The remaining 2 experiments have rarely or never been tested. All experiments have been performed using 10-fold cross validation.

### 5.1. Experiment 1: Normal vs Ictal vs Interictal Classification (AB vs CD vs E)

Zhang et al. (T. Zhang et al., 2017) pointed out that almost 100% accuracy has been achieved by several recent research works for normal vs epileptic or non-seizure vs seizure EEG signals classification. However, less work has been devoted to normal vs interictal vs ictal signals classification. They proposed a system targeting specifically this three-class problem, which achieved an accuracy of 97.35%.

Using M5 model, we achieved a mean accuracy of 96.1% with single P-1D-CNN model and 99.1% with an ensemble of 3 P-1D-CNN models, outperforming (T. Zhang et al., 2017) by 1.7%. The detailed analysis of the performance for this problem is given in Table. 3, which shows the average results for all the models and the augmentation schemes. However, Table 4 and 5 show the 10-fold cross-validation results and the confusion matrix for this problem. Table 5 indicates that the main confusion arises between normal and inter-ictal or inter-ictal and ictal.

**Table 4**. The accuracies of three-class problem (AB vs CD vs E) using model M5 and 10-fold cross validation

| Fold | $K1$ | $K2$ | $K3$ | $K4$ | $K5$ | $K6$ | $K7$ | $K8$ | $K9$ | $K10$ | Mean Acc |
|---|---|---|---|---|---|---|---|---|---|---|---|
| $Acc$ | 96.5 | 97.2 | 96.7 | 94.7 | 97.8 | 96.7 | 95.5 | 96.2 | 96 | 93.8 | 96.1 |
| $Acc_V$ | 99 | 100 | 99 | 97 | 100 | 100 | 100 | 98 | 99 | 99 | 99.1 |

**Table 5.** Confusion matrix for the three-class problem (AB vs CD vs E) using model M5, the values are the mean numbers of classified signals in 10-folds.

| | Normal (AB) | Interictal (CD) | Ictal (E) |
|---|---|---|---|
| Normal (AB) | 234 | 5 | 1 |
| Interictal (CD) | 23 | 217 | 0 |
| Ictal (E) | 0 | 5 | 115 |

### 5.2. Experiment 2: Normal vs Epileptic Classification (AB vs CDE and AB vs CD)

This case involves two types of experiments involving binary classification problems: (i) normal (AB) vs non-seizure epileptic (CD), and (ii) normal (AB) vs non-seizure and seizure epileptic (CDE); the 10-fold cross-validation results are shown in Table. 6. The mean accuracy of the proposed system for AB vs CD is 98.2% with single P-1D-CNN model, while 99.8% with the ensemble of 3 P-1D-CNN models. Similarly, the mean sensitivity and specificity are 98% and 99%, respectively. In the case of AB vs CDE, the mean accuracies are 98.1% and 99.95% with single model and ensemble, respectively, whereas both the mean sensitivity and specificity are 98%. The results indicate that the proposed system has better generalization and outperforms the state-of-the-art method reported in (Sharma et al., 2017; Sharmila et al., 2016). Also, it points out that ensemble of P-1D-CNN models performs better than single P-1D-CNN model, the reason is that in ensemble each model works an expert which analyses a local part of the signal, and finally local decisions are fused using majority vote to take the final decision.

**Table 6.** Performance Results of Normal vs Epileptic case using model $M5$ with 10-fold cross validation, here $Ki$ means $i$th fold.

| | | $K1$ | $K2$ | $K3$ | $K4$ | $K5$ | $K6$ | $K7$ | $K8$ | $K9$ | $K10$ | Mean |
|---|---|---|---|---|---|---|---|---|---|---|---|---|
| AB vs CD | $Acc$ | 99.2 | 96.3 | 98.8 | 97.9 | 98.3 | 99.6 | 97.3 | 99.2 | 96.9 | 98.5 | 98.2 |
| | $Sen$ | 100 | 95 | 98 | 97 | 98 | 100 | 95 | 100 | 96 | 99 | 98 |
| | $Spe$ | 98 | 97 | 100 | 99 | 98 | 99 | 99 | 99 | 98 | 98 | 99 |
| | $Acc_V$ | 100 | 100 | 99.4 | 100 | 100 | 100 | 98.8 | 100 | 99.4 | 100 | **99.8** |
| | | | | | | | | | | | | |
| AB vs CDE | $Acc$ | 95.8 | 99.3 | 98.5 | 99.7 | 96.8 | 96.7 | 98.7 | 99.2 | 97.8 | 98.2 | 98.1 |
| | $Sen$ | 98 | 99 | 99 | 100 | 96 | 99 | 99 | 99 | 98 | 97 | 98 |
| | $Spe$ | 92 | 100 | 98 | 99 | 98 | 93 | 98 | 99 | 98 | 100 | 98 |
| | $Acc_V$ | 99.5 | 100 | 100 | 100 | 100 | 100 | 100 | 100 | 100 | 100 | **99.95** |

### 5.3. Experiment 3: Normal or Non-Seizure vs Seizure Classification (A vs E, B vs E, A+B vs E, C vs E, D vs E, C+D vs E))

Third set of experiments involve six binary class problems ( (i) normal (A) vs seizure (E), (ii) normal (B) vs seizure (E), (iii) normal (AB) vs seizure (E), (iv) non- seizure (C) vs seizure (E), (v) non- seizure (D) vs seizure (E), and (vi) non- seizure (CD) vs seizure (E)). We tested all these combinations in order to check the power of the proposed system. Table. 7 reports the results. The mean accuracies given by single P-1D-CNN model varies from 99.9% to 97.4, whereas those given by ensemble varies from 100% to 98.5% for all the above problems. For all normal vs seizure problems, the accuracy is almost 100% with ensemble. For the problem C vs E, mean accuracy is 98.1% with single P-1D-CNN model and 98.5% with ensemble; in this case, there is a little improvement with ensemble, it indicates that in this case, almost all experts (P-1D-CNN models) have the save the same decision and it does not have a significant impact. For other two non-seizure vs seizure problems, the mean accuracies are 99.3% and 99.7%, which show that these are relatively easier problems than C vs E.

**Table 7.** Accuracies of 10-folds for Normal or Seizure vs Non-Seizure using model $M5$.

| | | $K1$ | $K2$ | $K3$ | $K4$ | $K5$ | $K6$ | $K7$ | $K8$ | $K9$ | $K10$ | Mean |
|---|---|---|---|---|---|---|---|---|---|---|---|---|
| A vs E | $Acc$ | 99.2 | 100 | 100 | 100 | 100 | 100 | 100 | 99.6 | 100 | 100 | 99.9 |
| | $Acc_V$ | 100 | 100 | 100 | 100 | 100 | 100 | 100 | 100 | 100 | 100 | 100 |
| | | | | | | | | | | | | |
| B vs E | $Acc$ | 100 | 95.4 | 98.8 | 100 | 97.1 | 98.8 | 100 | 100 | 100 | 100 | 99 |
| | $Acc_V$ | 100 | 97.5 | 100 | 100 | 98.8 | 100 | 100 | 100 | 100 | 100 | 99.6 |
| | | | | | | | | | | | | |
| AB vs E | $Acc$ | 99.2 | 96.4 | 97.5 | 98.9 | 96.7 | 100 | 100 | 99.7 | 100 | 100 | 98.8 |
| | $Acc_V$ | 100 | 99.2 | 98.3 | 100 | 99.2 | 100 | 100 | 100 | 100 | 100 | 99.7 |
| | | | | | | | | | | | | |
| C vs E | $Acc$ | 95 | 99.2 | 99.6 | 92.9 | 100 | 99.2 | 100 | 100 | 97.9 | 97.1 | 98.1 |
| | $Acc_V$ | 95 | 98.8 | 100 | 93.8 | 100 | 100 | 100 | 100 | 100 | 97.5 | 98.5 |
| | | | | | | | | | | | | |
| D vs E | $Acc$ | 98.8 | 100 | 97.5 | 99.6 | 93.8 | 94.6 | 97.9 | 97.9 | 98.8 | 95 | 97.4 |
| | $Acc_V$ | 100 | 100 | 100 | 100 | 98.8 | 98.8 | 100 | 100 | 100 | 95 | 99.3 |
| | | | | | | | | | | | | |
| CD vs E | $Acc$ | 99.4 | 98.9 | 100 | 99.4 | 97.5 | 100 | 99.2 | 98.9 | 98.1 | 96.1 | 98.8 |
| | $Acc_V$ | 100 | 100 | 100 | 100 | 100 | 100 | 100 | 100 | 100 | 96.7 | 99.7 |

### 5.4. Comparison with State-of-the-art Methods

Many methods have been proposed for the classification of EEG signals in binary (Normal vs Epileptic and seizure vs non-seizure) and ternary (Normal vs Interictal vs Ictal) classification problems. A comparison with state-of-the-art methods is given in Table 8; Zhang-17 ((T. Zhang et al., 2017), Sharma-17 (Sharma et al., 2017), Swami-16 (Swami et al., 2016), Sharmila-16 (Sharmila et al., 2016), Samiee-15 (Samiee et al., 2015), Orhan-11 (Orhan et al, 2011), Tzallas-12 (Tzallas et al., 2012). According to our knowledge until this date, DL approach has never been used for this problem. Recently, a fusion technique using variational mode decomposition (VMD) and an auto-regression based quadratic feature extraction technique have been proposed in (T. Zhang et al., 2017). Random forest classifier has been used to classify the extracted features into three categories. Despite using multiple complex techniques, it achieved 97.35% accuracy for three class problem, our system achieves 1.7% higher accuracy i.e. 99.1%.

The method proposed in (Sharmila et al., 2016) employ discrete wavelet transform (DWT) for feature extraction and nonlinear classifiers i.e. naïve Bayes (NB) and k-nearest neighbor (k-NN) classifier for the classification of epileptic and non-epileptic signals. The results reported for this method are without 10-fold cross validation. In spite of this fact, as is shown in Table 8, overall the proposed system gives the 10-fold cross validation results which outperform those reported in (Sharmila et al., 2016). This shows the robustness of the proposed system based on ensemble of deep P-1D-CNN models and indicates that it has better generalization than state-of-the-art methods. The mean accuracy of the proposed system is 99.6% for all the sixteen cases (shown in Table 8 last column), which figures out the generalization power of the proposed system.

**Table. 8.** Performance of model H with Case 3 window sliding on all combination of binary and ternary class classification scheme. Some of the abbreviation used are Time frequency features (TF)

| Data Sets Combination | Methodology | 10-fold CV | Stat-of-the-Art | Acc | Our Acc |
|---|---|---|---|---|---|
| AB vs CD vs E | VMD+AR+RF | Yes | Zhang-17 | 97.4 | **99.1** |
| AB vs CD | ATFFWT + LS-SVM | Yes | Sharma-17 | 92.5 | **99.9** |
| AB vs E | ATFFWT + LS-SVM | Yes | Sharma-17 | **100** | 99.8 |
| | DTCWT + GRNN | Yes | Swami-16 | 99.2 | |
| A vs E | DWT+NB/K-NN | No | Sharmila-16 | 100 | **100** |
| | TF + ANN | Yes | Tzallas-12 | 100 | |
| | DTCWT + GRNN | Yes | Swami-16 | 100 | |
| B vs E | ATFFWT + LS-SVM | Yes | Sharma-17 | **100** | 99.8 |
| | DTCWT + GRNN | Yes | Swami-16 | 98.9 | |
| CD vs E | DWT+NB/K-NN | No | Sharmila-16 | 98.8 | **99.7** |
| | ATFFWT + LS-SVM | Yes | Sharma-17 | 98.7 | |
| | DTCWT + GRNN | Yes | Swami-16 | 95.2 | |
| C vs E | ATFFWT + LS-SVM | Yes | Sharma-17 | 99 | **99.1** |
| | DTCWT + GRNN | Yes | Swami-16 | 98.7 | |
| D vs E | ATFFWT + LS-SVM | Yes | Sharma-17 | 98.5 | **99.4** |
| | DTCWT + GRNN | Yes | Swami-16 | 93.3 | |
| BCD vs E | DWT+NB/K-NN | No | Sharmila-16 | 96.4 | **99.3** |
| BC vs E | DWT+NB/K-NN | No | Sharmila-16 | 98.3 | **99.5** |
| BD vs E | DWT+NB/K-NN | No | Sharmila-16 | 96.5 | **99.6** |
| AC vs E | DWT+NB/K-NN | No | Sharmila-16 | 99.6 | **99.7** |
| ABCD vs E | ATFFWT + LS-SVM | Yes | Sharma-17 | 99.2 | **99.7** |
| | TF + ANN | Yes | Tzallas-12 | 97.7 | |
| | DWT+MLP | No | Orhan-11 | 99.6 | |
| | FT+MLP | No | Samiee-15 | 98.1 | |
| | DTCWT + GRNN | Yes | Swami-16 | 95.24 | |
| AB vs CDE | DWT+NB/K-NN | No | Sharmila-16 | - | **99.5** |
| ABC vs E | DWT+NB/K-NN | No | Sharmila-16 | 98.68 | **99.97** |
| ACD vs E | DWT+NB/K-NN | No | Sharmila-16 | 97.31 | **99.8** |

## 6. Conclusion

In this paper, an automatic system for epilepsy detection has been proposed, which deals with binary detection problems (epileptic vs. non-epileptic or seizure vs. non-seizure) and ternary detection problem (ictal vs. normal vs. interictal). The proposed system is based on deep learning, which is state-of-the-art ML approach. For this system, a memory efficient and simple pyramidal one-dimensional deep convolutional neural network (P-1D-CNN) model has been introduced, which is an end-to-end model, and involves less number of learnable parameters. The system has been designed as an ensemble of P-1D-CNN models, which takes an EEG signal as input, passes it to different P-1D-CNN models and finally fuses their decisions using majority vote. To overcome the issue of small dataset, two data augmentation schemes have been introduced for learning P-1D-CNN model. Due to fewer parameters, P-1D-CNN model is easy to train as well as easy to deploy on chips where memory is limited. The proposed system gives outstanding performance with less data and fewer parameters. It will assist neurologists in detecting epilepsy, and will greatly reduce their burden and increase their efficiency. In almost all the cases concerning epilepsy detection, the proposed system gives an accuracy of 99.1±0.9% on the University of Bonn dataset. The system can be useful for other similar classification problems based on EEG brain signals. Currently, the epilepsy detection methods detect seizures after their occurrence. In future, we will investigate its usefulness for detecting seizures prior to their occurrence, which is a challenging problem.